# IDL-Expressions: A Formalism for Representing and Parsing Finite Languages in Natural Language Processing

**Mark-Jan Nederhof** MARKJAN@LET.RUG.NL
*Faculty of Arts, University of Groningen*
*P.O. Box 716*
*NL-9700 AS Groningen, The Netherlands*

**Giorgio Satta** SATTA@DEI.UNIPD.IT
*Dept. of Information Engineering, University of Padua*
*via Gradenigo, 6/A*
*I-35131 Padova, Italy*

## Abstract

We propose a formalism for representation of finite languages, referred to as the class of *IDL-expressions*, which combines concepts that were only considered in isolation in existing formalisms. The suggested applications are in natural language processing, more specifically in surface natural language generation and in machine translation, where a sentence is obtained by first generating a large set of candidate sentences, represented in a compact way, and then filtering such a set through a parser. We study several formal properties of IDL-expressions and compare this new formalism with more standard ones. We also present a novel parsing algorithm for IDL-expressions and prove a non-trivial upper bound on its time complexity.

## 1. Introduction

In natural language processing, more specifically in applications that involve natural language generation, the task of *surface generation* consists in the process of generating an output sentence in a target language, on the basis of some input representation of the desired meaning for the output sentence. During the last decade, a number of new approaches for natural language surface generation have been put forward, called *hybrid approaches*. Hybrid approaches make use of symbolic knowledge in combination with statistical techniques that have recently been developed for natural language processing. Hybrid approaches therefore share many advantages with statistical methods for natural language processing, such as high accuracy, wide coverage, robustness, portability and scalability.

Hybrid approaches are typically based on two processing phases, described in what follows (Knight & Hatzivassiloglou, 1995; Langkilde & Knight, 1998; Bangalore & Rambow, 2000 report examples of applications of this approach in real world generation systems). In the first phase one generates a large set of candidate sentences by a relatively simple process. This is done on the basis of an input sentence in some source language in case the process is embedded within a machine translation system, or more generally on the basis of some logical/semantic representation, called *conceptual structure*, which denotes the meaning that the output sentence should convey. This first phase involves no or only few intricacies of the target language, and the set of candidate sentences may contain many that are ungrammatical or that can otherwise be seen as less desirable than others. In the second phase one or more preferred sentences are selected from the collection of candidates, exploiting some form of syntactic processing that more heavily relies on properties of the target language than the first phase. This syntactic processing may involve language models as simple as bigrams or it may involve more powerful models such as those based on context-free grammars, which typically perform with higher accuracy on this task (see for instance work presented by Charniak, 2001 and references therein).

In hybrid approaches, the generation of the candidate set typically involves a symbolic grammar that has been quickly hand-written, and is quite small and easy to maintain. Such a grammar cannot therefore account for all of the intricacies of the target language. For instance, frequency information for synonyms and collocation information in general is not encoded in the grammar. Similarly, lexico-syntactic and selectional constraints for the target language might not be fully specified, as is usually the case with small and mid-sized grammars. Furthermore, there might also be some underspecification stemming from the input conceptual structure. This is usually the case if the surface generation module is embedded into a larger architecture for machine translation, and the source language is underspecified for features such as definiteness, time and number. Since inferring the missing information from the sentence context is a very difficult task, the surface generation module usually has to deal with underspecified knowledge.

All of the above-mentioned problems are well-known in the literature on natural language surface generation, and are usually referred to as "lack of knowledge" of the system or of the input. As a consequence of these problems, the set of candidate sentences generated in the first phase may be extremely large. In real world generation systems, candidate sets have been reported to contain as many as $10^{12}$ sentences (Langkilde, 2000). As already explained, the second processing phase in hybrid approaches is intended to reduce these huge sets to subsets containing only a few sentences. This is done by exploiting knowledge about the target language that was not available in the first phase. This additional knowledge can often be obtained through automatic extraction from corpora, which requires considerably less effort than the development of hand-written, purely symbolic systems.

Due to the extremely large size of the set of candidate sentences, the feasibility of hybrid approaches to surface natural language generation relies on

- the compactness of the representation of a set of candidate sentences that in real world systems might be as large as $10^{12}$; and
- the efficiency of syntactic processing of the stored set.

Several solutions have been adopted in existing hybrid systems for the representation of the set of candidate sentences. These include bags of words (Brown et al., 1990) and bags of complex lexical representations (Beaven, 1992; Brew, 1992; Whitelock, 1992), word lattices (Knight & Hatzivassiloglou, 1995; Langkilde & Knight, 1998; Bangalore & Rambow, 2000), and non-recursive context-free grammars (Langkilde, 2000). As will be discussed in detail in Section 2, word lattices and non-recursive context-free grammars allow encoding of precedence constraints and choice among different words, but they both lack a primitive for representing strings that are realized by combining a collection of words in an arbitrary order. On the other hand, bags of words allow the encoding of free word order, but in such a representation one cannot directly express precedence constraints and choice among different words.

In this paper we propose a new representation that combines all of the above-mentioned primitives. This representation consists of *IDL-expressions*. In the term IDL-expression, 'I' stands for "interleave", which pertains to phrases that may occur interleaved, allowing freedom on word order (a precise definition of this notion will be provided in the next section); 'D' stands for "disjunction", which allows choices of words or phrases; 'L' stands for "lock", which is used to constrain the application of the interleave operator. We study some interesting properties of this representation, and argue that the expressivity of the formalism makes it more suitable than the alternatives discussed above for use within hybrid architectures for surface natural language generation. We also associate IDL-expressions with *IDL-graphs*, an equivalent representation that can be more easily interpreted by a machine, and develop a dynamic programming algorithm for parsing IDL-graphs using a context-free grammar. If a set of candidate sentences is represented as an IDL-expression or IDL-graph, the algorithm can be used to filter out ungrammatical sentences from the set, or to rank the sentences in the set according to their likelihood, in case the context-free grammar assigns weights to derivations. While parsing is traditionally defined for input consisting of a single string, we here conceive parsing as a process that can be carried out on an input device denoting a language, i.e., a set of strings.

There is a superficial similarity between the problem described above of representing finite sets in surface generation, and a different research topic, often referred to as discontinuous parsing. In discontinuous parsing one seeks to relax the definition of context-free grammars in order to represent the syntax of languages that exhibit constructions with uncertainty on word or constituent order (see for instance work reported by Daniels & Meurers, 2002 and references therein). In fact, some of the operators we use in IDL-expressions have also been exploited in recent work on discontinuous parsing. However, the parsing problem for discontinuous grammars and the parsing problem for IDL-expressions are quite different: in the former, we are given a grammar with productions that express uncertainty on constituent order, and need to parse an input string whose symbols are totally ordered; in the latter problem we are given a grammar with total order on the constituents appearing in each production, and need to parse an input that includes uncertainty on word and constituent order.

This paper is structured as follows. In Section 2 we give a brief overview of existing representations of finite languages that have been used in surface generation components. We then discuss some notational preliminaries in Section 3. In Section 4 we introduce IDL-expressions and define their semantics. In Section 5 we associate with IDL-expressions an equivalent but more procedural representation, called IDL-graphs. We also introduce the important notion of cut of an IDL-graph, which will be exploited later by our algorithm. In Section 6 we briefly discuss the Earley algorithm, a traditional method for parsing a string using a context-free grammar, and adapt this algorithm to work on finite languages encoded by IDL-graphs. In Section 7 we prove a non-trivial upper bound on the number of cuts in an IDL-graph, and on this basis we investigate the computational complexity of our parsing algorithm. We also address some implementational issues. We conclude with some discussion in Section 8.

## 2. Representations of Finite Languages

In this section we analyze and compare existing representations of finite languages that have been adopted in surface generation components of natural language systems.

Bags (or multisets) of words have been used in several approaches to surface generation. They are at the basis of the generation component of the statistical machine translation models proposed by Brown et al. (1990). Bags of complex lexical signs have also been used in the machine translation approach described by Beaven (1992) and Whitelock (1992), called shake-and-bake. As already mentioned, bags are a very succinct representation of finite languages, since they allow encoding of more than exponentially many strings in the size of a bag itself. This power comes at a cost, however. Deciding whether some string encoded by an input bag can be parsed by a CFG is NP-complete (Brew, 1992). It is not difficult to show that this result still holds in the case of a regular grammar or, equivalently, a regular expression. An NP-completeness result involving bags has also been presented by Knight (1999), for a related problem where the parsing grammar is a probabilistic model based on bigrams.

As far as expressivity is concerned, bags of words have also strict limitations. These structures lack a primitive for expressing choices among words. As already observed in the introduction, this is a serious problem in natural language generation, where alternatives in lexical realization must be encoded in the presence of lack of detailed knowledge of the target language. In addition, bags of words usually do not come with precedence constraints. However, in natural language applications these constraints are very common, and are usually derived from knowledge about the target language or, in the case of machine translation, from the parsing tree of the source string. In order to represent these constraints, extra machinery must be introduced. For instance, Brown et al. (1990) impose, for each word in the bag, a probabilistic distribution delimiting its position in the target string, on the basis of the original position of the source word in the input string to be translated. In the shake and bake approach, bags are defined over functional structures, each representing complex lexical information from which constraints can be derived. Then the parsing algorithm for bags is interleaved with a constraint propagation algorithm to filter out parses (e.g., as done by Brew, 1992). As a general remark, having different layers of representation requires the development of more involved parsing algorithms, which we try to avoid in the new proposal to be described below.

An alternative representation of finite languages is the class of acyclic deterministic finite automata, also called word lattices. This representation has often been used in hybrid approaches to surface generation (Knight & Hatzivassiloglou, 1995; Langkilde & Knight, 1998; Bangalore & Rambow, 2000), and more generally in natural language applications where some form of uncertainty comes with the input, as for instance in speech recognition (Jurafsky & Martin, 2000, Section 7.4). Word lattices inherit from standard regular expressions the primitives expressing concatenation and disjunction, and thereby allow the encoding of precedence constraints and word disjunction in a direct way. Furthermore, word lattices can be efficiently parsed by means of CFGs, using standard techniques for lattice parsing (Aust, Oerder, Seide, & Steinbiss, 1995). Lattice parsing requires cubic time in the number of states of the input finite automaton and linear time in the size of the CFG. Methods for lattice parsing can all be traced back to Bar-Hillel, Perles, and Shamir (1964), who prove that the class of context-free languages is closed under intersection with regular languages.

One limitation of word lattices and finite automata in general is the lack of an operator for free word order. As we have already discussed in the introduction, this is a severe limitation for hybrid systems, where free word order in sentence realization is needed in case the symbolic grammar used in the first phase fails to provide ordering constraints. To represent strings where a bag of words can occur in every possible order, one has to encode each string through an individual path within the lattice. In the general case, this requires an amount of space that is more than exponential in the size of the bag. From this perspective, the previously mentioned polynomial time result for parsing is to no avail, since the input structure to the parser might already be of size more than exponential in the size of the input conceptual structure. The problem of free word order in lattice structures is partially solved by Langkilde and Knight (1998) by introducing an external recasting mechanism that preprocesses the input conceptual structure. This has the overall effect that phrases normally represented by two independent sublattices can now be generated one embedded into the other, therefore partially mimicking the interleaving of the words in the two phrases. However, this is not enough to treat free word order in its full generality.

A third representation of finite languages, often found in the literature on compression theory (Nevill-Manning & Witten, 1997), is the class of non-recursive CFGs. A CFG is called *non-recursive* if no nonterminal can be rewritten into a string containing the nonterminal itself. It is not difficult to see that such grammars can only generate finite languages. Non-recursive CFGs have recently been exploited in hybrid systems (Langkilde, 2000).[1] This representation inherits all the expressivity of word lattices, and thus can encode precedence constraints as well as disjunctions. In addition, non-recursive CFGs can achieve much smaller encodings of finite languages than word lattices. This is done by uniquely encoding certain sets of substrings that occur repeatedly through a nonterminal that can be reused in several places. This feature turns out to be very useful for natural language applications, as shown by experimental results reported by Langkilde (2000).

Although non-recursive CFGs can be more compact representations than word lattices, this representation still lacks a primitive for representing free word order. In fact, a CFG generating the finite language of all permutations of $n$ symbols must have size at least exponential in $n$.[2] In addition, the problem of deciding whether some string encoded by a non-recursive CFG can be parsed by a general CFG is PSPACE-complete (Nederhof & Satta, 2004).

From the above discussion, one can draw the following conclusions. In considering the range of possible encodings for finite languages, we are interested in measuring (i) the compactness of the representation, and (ii) the efficiency of parsing the obtained representation by means of a CFG. At one extreme we have the naive solution of enumerating all strings in the language, and then independently parsing each individual string using a traditional string parsing algorithm. This solution is obviously unfeasible, since no compression at all is achieved and so the overall amount of time required might be exponential in the size of

1. Langkilde (2000) uses the term “forests” for non-recursive CFGs, which is a different name for the same concept (Billot & Lang, 1989).
2. An unpublished proof of this fact has been personally communicated to the authors by Jeffrey Shallit and Ming-wei Wang.

the input conceptual structure. Although word lattices are a more compact representation, when free word order needs to be encoded we may still have representations of exponential size as input to the parser, as already discussed. At the opposite extreme, we have solutions like bags of words or non-recursive CFGs, which allow very compact representations, but are still very demanding in parsing time requirements. Intuitively, this can be explained by considering that parsing a highly compressed finite language requires additional bookkeeping with respect to the string case. What we then need to explore is some trade-off between these solutions, offering interesting compression factors at the expense of parsing time requirements that are provably polynomial in the cases of interest. As we will show in the sequel of this paper, IDL-expressions have these required properties and are therefore an interesting solution to the problem.

## 3. Notation

In this section we briefly recall some basic notions from formal language theory. For more details we refer the reader to standard textbooks (e.g., Harrison, 1978).

For a set $\Delta$, $|\Delta|$ denotes the number of elements in $\Delta$; for a string $x$ over some alphabet, $|x|$ denotes the length of $x$. For string $x$ and languages (sets of strings) $L$ and $L'$, we let $x \cdot L = \{xy \mid y \in L\}$ and $L \cdot L' = \{xy \mid x \in L, y \in L'\}$. We remind the reader that a string-valued function $f$ over some alphabet $\Sigma$ can be extended to a homomorphism over $\Sigma^*$ by letting $f(\varepsilon) = \varepsilon$ and $f(ax) = f(a)f(x)$ for $a \in \Sigma$ and $x \in \Sigma^*$. We also let $f(L) = \{f(x) \mid x \in L\}$.

We denote a *context-free grammar* (CFG) by a 4-tuple $G = (N, \Sigma, P, S)$, where $N$ is a finite set of nonterminals, $\Sigma$ is a finite set of terminals, with $\Sigma \cap N = \emptyset$, $S \in N$ is a special symbol called the start symbol, and $P$ is a finite set of productions having the form $A \rightarrow \gamma$, with $A \in N$ and $\gamma \in (\Sigma \cup N)^*$. Throughout the paper we assume the following conventions: $A, B, C$ denote nonterminals, $a, b, c$ denote terminals, $\alpha, \beta, \gamma, \delta$ denote strings in $(\Sigma \cup N)^*$ and $x, y, z$ denote strings in $\Sigma^*$.

The derives relation is denoted $\Rightarrow_G$ and its transitive closure $\Rightarrow_G^+$. The language generated by grammar $G$ is denoted $L(G)$. The size of $G$ is defined as

$$|G| = \sum_{(A \rightarrow \alpha) \in P} |A\alpha| . \tag{1}$$

## 4. IDL-Expressions

In this section we introduce the class of IDL-expressions and define a mapping from such expressions to sets of strings. Similarly to regular expressions, IDL-expressions generate sets of strings, i.e., languages. However, these languages are always finite. Therefore the class of languages generated by IDL-expressions is a proper subset of the class of regular languages. As already discussed in the introduction, IDL-expressions combine language operators that were only considered in isolation in previous representations of finite languages exploited in surface natural language generation. In addition, some of these operations have been recently used in the discontinuous parsing literature, for the syntactic description of (infinite) languages with weak linear precedence constraints. IDL-expressions represent choices among words or phrases and their relative ordering by means of the standard concatenation operator '·' from regular expressions, along with three additional operators to be discussed in what follows. All these operators take as arguments one or more IDL-expressions, and combine the strings generated by these arguments in different ways.

- Operator '‖', called **interleave**, interleaves strings resulting from its argument expressions. A string $z$ results from the interleaving of two strings $x$ and $y$ whenever $z$ is composed of all and only the occurrences of symbols in $x$ and $y$, and these symbols appear within $z$ in the same relative order as within $x$ and $y$. As an example, consider strings `abcd` and `efg`. By interleaving these two strings we obtain, among many others, the strings `abecfgd`, `eabfgcd` and `efabcdg`. In the formal language literature, this operation has also been called 'shuffle', as for instance by Dassow and Păun (1989). In the discontinuous parsing literature and in the literature on head-driven phrase-structure grammars (HPSG, Pollard & Sag, 1994) the interleave operation is also called 'sequence union' (Reape, 1989) or 'domain union' (Reape, 1994). The interleave operator also occurs in an XML tool described by van der Vlist (2003).

- Operator '∨', called **disjunction**, allows a choice between strings resulting from its argument expressions. This is a standard operator from regular expressions, where it is more commonly written as '+'.

- Operator '×', called **lock**, takes a single IDL-expression as argument. This operator states that no additional material can be interleaved with a string resulting from its argument. The lock operator has been previously used in the discontinuous parsing literature, as for instance by Daniels and Meurers (2002), Götz and Penn (1997), Ramsay (1999), Suhre (1999). In that context, the operator was called 'isolation'.

The interleave, disjunction and lock operators will also be called I, D and L operators, respectively. As we will see later, the combination of the I and L operators within IDL-expressions provides much of the power of existing formalisms to represent free word order, while maintaining computational properties quite close to those of regular expressions or finite automata.

As an introductory example, we discuss the following IDL-expression, defined over the word alphabet {`piano`, `play`, `must`, `necessarily`, `we`}.

$$\|(\vee(\texttt{necessarily}, \texttt{must}),\ \texttt{we} \cdot \times(\texttt{play} \cdot \texttt{piano})). \tag{2}$$

IDL-expression (2) says that words `we`, `play` and `piano` must appear in that order in any of the generated strings, as specified by the two occurrences of the concatenation operator. Furthermore, the use of the lock operator states that no additional words can ever appear in between `play` and `piano`. The disjunction operator expresses the choice between words `necessarily` and `must`. Finally, the interleave operator states that the word resulting from the first of its arguments must be inserted into the sequence `we`, `play`, `piano`, in any of the available positions. Notice the interaction with the lock operator, which, as we have seen, makes unavailable the position in between `play` and `piano`. Thus the following sentences, among others, can be generated by IDL-expression (2):

`necessarily we play piano`
`must we play piano`
`we must play piano`
`we play piano necessarily.`

However, the following sentences cannot be generated by IDL-expression (2):

`we play necessarily piano`
`necessarily must we play piano.`

The first sentence is disallowed through the use of the lock operator, and the second sentence is impossible because the disjunction operator states that exactly one of the arguments must appear in the sentence realization. We now provide a formal definition of the class of IDL-expressions.

**Definition 1** *Let $\Sigma$ be some finite alphabet and let $\mathcal{E}$ be a symbol not in $\Sigma$. An* **IDL-expression** *over $\Sigma$ is a string $\pi$ satisfying one of the following conditions:*

(i) $\pi = a$, *with* $a \in \Sigma \cup \{\mathcal{E}\}$;

(ii) $\pi = \times(\pi')$, *with $\pi'$ an IDL-expression;*

(iii) $\pi = \vee(\pi_1, \pi_2, \ldots, \pi_n)$, *with $n \geq 2$ and $\pi_i$ an IDL-expression for each $i$, $1 \leq i \leq n$;*

(iv) $\pi = \|(\pi_1, \pi_2, \ldots, \pi_n)$, *with $n \geq 2$ and $\pi_i$ an IDL-expression for each $i$, $1 \leq i \leq n$;*

(v) $\pi = \pi_1 \cdot \pi_2$, *with $\pi_1$ and $\pi_2$ both IDL-expressions.*

We take the infix operator '$\cdot$' to be right associative, although in all of the definitions in this paper, disambiguation of associativity is not relevant and can be taken arbitrarily. We say that IDL-expression $\pi'$ is a subexpression of $\pi$ if $\pi'$ appears as an argument of some operator in $\pi$.

We now develop a precise semantics for IDL-expressions. The only technical difficulty in doing so arises with the proper treatment of the lock operator.[3] Let $x$ be a string over $\Sigma$. The basic idea below is to use a new symbol $\diamond$, not already in $\Sigma$. An occurrence of $\diamond$ between two terminals indicates that an additional string can be inserted at that position. As an example, if $x = x' \diamond x''x'''$ with $x'$, $x''$ and $x'''$ strings over $\Sigma$, and if we need to interleave $x$ with a string $y$, then we may get as a result string $x'yx''x'''$ but not string $x' \diamond x''yx'''$. The lock operator corresponds to the removal of every occurrence of $\diamond$ from a string.

More precisely, strings in $(\Sigma \cup \{\diamond\})^*$ will be used to represent sequences of strings over $\Sigma$; symbol $\diamond$ is used to separate the strings in the sequence. Furthermore, we introduce a string homomorphism $\mathsf{lock}$ over $(\Sigma \cup \{\diamond\})^*$ by letting $\mathsf{lock}(a) = a$ for $a \in \Sigma$ and $\mathsf{lock}(\diamond) = \epsilon$. An application of $\mathsf{lock}$ to an input sequence can be seen as the operation of concatenating together all of the strings in the sequence.

3. If we were to add the Kleene star, then infinite languages can be specified, and interleave and lock can be more conveniently defined using derivatives (Brzozowski, 1964), as noted before by van der Vlist (2003).

We can now define the basic operation $\mathsf{comb}$, which plays an important role in the sequel. This operation composes two sequences $x$ and $y$ of strings, represented as explained above, into a set of new sequences of strings. This is done by interleaving the two input sequences in every possible way. Operation $\mathsf{comb}$ makes use of an auxiliary operation $\mathsf{comb}'$, which also constructs interleaved sequences out of input sequences $x$ and $y$, but always starting with the first string in its first argument $x$. As any sequence in $\mathsf{comb}(x,y)$ must start with a string from $x$ or with a string from $y$, $\mathsf{comb}(x,y)$ is the union of $\mathsf{comb}'(x,y)$ and $\mathsf{comb}'(y,x)$. In the definition of $\mathsf{comb}'$, we distinguish the case in which $x$ consists of a single string and the case in which $x$ consists of at least two strings. In the latter case, the tail of an output sequence can be obtained by applying $\mathsf{comb}$ recursively on the tail of sequence $x$ and the complete sequence $y$. For $x, y \in (\Sigma \cup \{\diamond\})^*$, we have:

$$\begin{aligned}
\mathsf{comb}(x,y) &= \mathsf{comb}'(x,y) \cup \mathsf{comb}'(y,x) \\
\mathsf{comb}'(x,y) &= \begin{cases} \{x \diamond y\}, & \text{if } x \in \Sigma^*; \\ \{x'\diamond\} \cdot \mathsf{comb}(x'', y), & \\ \quad \text{if there are } x' \in \Sigma^* \text{ and } x'' & \\ \quad \text{such that } x = x' \diamond x''. & \end{cases}
\end{aligned}$$

As an example, let $\Sigma = \{\mathsf{a}, \mathsf{b}, \mathsf{c}, \mathsf{d}, \mathsf{e}\}$ and consider the two sequences $\mathsf{a} \diamond \mathsf{bb} \diamond \mathsf{c}$ and $\mathsf{d} \diamond \mathsf{e}$. Then we have

$$\begin{aligned}
&\mathsf{comb}(\mathsf{a} \diamond \mathsf{bb} \diamond \mathsf{c}, \mathsf{d} \diamond \mathsf{e}) = \\
&\quad \{\mathsf{a} \diamond \mathsf{bb} \diamond \mathsf{c} \diamond \mathsf{d} \diamond \mathsf{e}, \mathsf{a} \diamond \mathsf{bb} \diamond \mathsf{d} \diamond \mathsf{c} \diamond \mathsf{e}, \mathsf{a} \diamond \mathsf{bb} \diamond \mathsf{d} \diamond \mathsf{e} \diamond \mathsf{c}, \\
&\quad \mathsf{a} \diamond \mathsf{d} \diamond \mathsf{bb} \diamond \mathsf{c} \diamond \mathsf{e}, \mathsf{a} \diamond \mathsf{d} \diamond \mathsf{bb} \diamond \mathsf{e} \diamond \mathsf{c}, \mathsf{a} \diamond \mathsf{d} \diamond \mathsf{e} \diamond \mathsf{bb} \diamond \mathsf{c}, \\
&\quad \mathsf{d} \diamond \mathsf{a} \diamond \mathsf{bb} \diamond \mathsf{c} \diamond \mathsf{e}, \mathsf{d} \diamond \mathsf{a} \diamond \mathsf{bb} \diamond \mathsf{e} \diamond \mathsf{c}, \mathsf{d} \diamond \mathsf{a} \diamond \mathsf{e} \diamond \mathsf{bb} \diamond \mathsf{c}, \\
&\quad \mathsf{d} \diamond \mathsf{e} \diamond \mathsf{a} \diamond \mathsf{bb} \diamond \mathsf{c}\}.
\end{aligned}$$

For languages $L_1, L_2$ we define $\mathsf{comb}(L_1, L_2) = \cup_{x \in L_1, y \in L_2} \mathsf{comb}(x,y)$. More generally, for languages $L_1, L_2, \ldots, L_d$, $d \geq 2$, we define $\mathsf{comb}_{i=1}^{d} L_i = \mathsf{comb}(L_1, L_2)$ for $d = 2$, and $\mathsf{comb}_{i=1}^{d} L_i = \mathsf{comb}(\mathsf{comb}_{i=1}^{d-1} L_i, L_d)$ for $d > 2$.

**Definition 2** *Let $\Sigma$ be some finite alphabet. Let $\sigma$ be a function mapping IDL-expressions over $\Sigma$ into subsets of $(\Sigma \cup \{\diamond\})^*$, specified by the following conditions:*

(i) $\sigma(a) = \{a\}$ *for* $a \in \Sigma$, *and* $\sigma(\mathcal{E}) = \{\varepsilon\}$;

(ii) $\sigma(\times(\pi)) = \mathsf{lock}(\sigma(\pi))$;

(iii) $\sigma(\vee(\pi_1, \pi_2, \ldots, \pi_n)) = \cup_{i=1}^{n} \sigma(\pi_i)$;

(iv) $\sigma(\|(\pi_1, \pi_2, \ldots, \pi_n)) = \mathsf{comb}_{i=1}^{n} \sigma(\pi_i)$;

(v) $\sigma(\pi \cdot \pi') = \sigma(\pi) \diamond \sigma(\pi')$.

*The set of strings that satisfy an IDL-expression $\pi$, written $L(\pi)$, is given by $L(\pi) = \mathsf{lock}(\sigma(\pi))$.*

As an example for the above definition, we show how the interleave operator can be used in an IDL-expression to denote the set of all strings realizing permutations of a given bag of symbols. Let $\Sigma = \{\mathtt{a}, \mathtt{b}, \mathtt{c}\}$. Consider a bag $\langle \mathtt{a}, \mathtt{a}, \mathtt{b}, \mathtt{c}, \mathtt{c} \rangle$ and IDL-expression

$$\|(\mathtt{a}, \mathtt{a}, \mathtt{b}, \mathtt{c}, \mathtt{c}). \tag{3}$$

By applying Definition 2 to IDL-expression (3), we obtain in the first few steps

$$\begin{aligned}
\sigma(\mathtt{a}) &= \{\mathtt{a}\}, \\
\sigma(\mathtt{b}) &= \{\mathtt{b}\}, \\
\sigma(\mathtt{c}) &= \{\mathtt{c}\}, \\
\sigma(\|(\mathtt{a}, \mathtt{a})) &= \mathtt{comb}(\{\mathtt{a}\}, \{\mathtt{a}\}) \;=\; \{\mathtt{a} \diamond \mathtt{a}\}, \\
\sigma(\|(\mathtt{a}, \mathtt{a}, \mathtt{b})) &= \mathtt{comb}(\{\mathtt{a} \diamond \mathtt{a}\}, \{\mathtt{b}\}) \;=\; \{\mathtt{b} \diamond \mathtt{a} \diamond \mathtt{a}, \mathtt{a} \diamond \mathtt{b} \diamond \mathtt{a}, \mathtt{a} \diamond \mathtt{a} \diamond \mathtt{b}\}.
\end{aligned}$$

In the next step we obtain $3 \times 4$ sequences of length 4, each using all the symbols from bag $\langle \mathtt{a}, \mathtt{a}, \mathtt{b}, \mathtt{c} \rangle$. One more application of the comb operator, on this set and set $\{\mathtt{c}\}$, provides all possible sequences of singleton strings expressing permutations of symbols in bag $\langle \mathtt{a}, \mathtt{a}, \mathtt{b}, \mathtt{c}, \mathtt{c} \rangle$. After removing symbol $\diamond$ throughout, which conceptually turns sequences of strings into undivided strings, we obtain the desired language $L(\|(\mathtt{a}, \mathtt{a}, \mathtt{b}, \mathtt{c}, \mathtt{c}))$ of permutations of bag $\langle \mathtt{a}, \mathtt{a}, \mathtt{b}, \mathtt{c}, \mathtt{c} \rangle$.

To conclude this section, we compare the expressivity of IDL-expressions with that of the formalisms discussed in Section 2. We do this by means of a simple example. In what follows, we use the alphabet $\{\mathtt{NP}, \mathtt{PP}, \mathtt{V}\}$. These symbols denote units standardly used in syntactic analysis of natural language, and stand for, respectively, noun phrase, prepositional phrase and verb. Symbols NP, PP and V should be rewritten into actual words of the language, but we use these as terminal symbols to simplify the presentation. Consider a language having the subject-verb-object (SVO) order and a sentence having the structure

$$[_{\mathtt{S}}\ \mathtt{NP}_1\ \mathtt{V}\ \mathtt{NP}_2],$$

where $\mathtt{NP}_1$ realizes the subject position and $\mathtt{NP}_2$ realizes the object position. Let $\mathtt{PP}_1$ and $\mathtt{PP}_2$ be phrases that must be inserted in the above sentence as modifiers. Assume that we know that the language at hand does not allow modifiers to appear in between the verbal and the object positions. Then we are left with 3 available positions for the realization of a first modifier, out of the 4 positions in our string. After the first modifier is inserted within the string, we have 5 positions, but only 4 are available for the realization of a second modifier, because of our assumption. This results in a total of $3 \times 4 = 12$ possible sentence realizations.

A bag of words for these sentences is unable to capture the above constraint on the positioning of modifiers. At the same time, a word lattice for these sentences would contain 12 distinct paths, corresponding to the different realizations of the modifiers in the basic sentence. Using the IDL formalism, we can easily capture the desired realizations by means of the IDL-expression:

$$\|(\mathtt{PP}_1, \mathtt{PP}_2, \mathtt{NP}_1 \cdot \times(\mathtt{V} \cdot \mathtt{NP}_2)).$$

Again, note the presence of the lock operator, which implements our restriction against modifiers appearing in between the verbal and the object position, similarly to what we have done in IDL-expression (2).

Consider now a sentence with a subordinate clause, having the structure

$$[_{\mathtt{S}}\ \mathtt{NP}_1\ \mathtt{V}_1\ \mathtt{NP}_2\ [_{\mathtt{S'}}\ \mathtt{NP}_3\ \mathtt{V}_2\ \mathtt{NP}_4]],$$

and assume that modifiers $\mathtt{PP}_1$ and $\mathtt{PP}_2$ apply to the main clause, while modifiers $\mathtt{PP}_3$ and $\mathtt{PP}_4$ apply to the subordinate clause. As before, we have $3 \times 4$ possible realizations for the subordinate sentence. If we allow main clause modifiers to appear in positions before the subordinate clause as well as after the subordinate clause, we have $4 \times 5$ possible realizations for the main sentence. Overall, this gives a total of $3 \times 4^2 \times 5 = 240$ possible sentence realizations.

Again, a bag representation for these sentences is unable to capture the above restrictions on word order, and would therefore badly overgenerate. Since the main sentence modifiers could be placed after the subordinate clause, we need to record for each of the two modifiers of the main clause whether it has already been seen, while processing the 12 possible realizations of the subordinate clause. This increases the size of the representation by a factor of $2 \times 2 = 4$. On the other hand, the desired realizations can be easily captured by means of the IDL-expression:

$$\|(\mathrm{PP}_1, \mathrm{PP}_2, \mathrm{NP}_1 \cdot \times(\mathrm{V}_1 \cdot \mathrm{NP}_2) \cdot \times(\|(\mathrm{PP}_3, \mathrm{PP}_4, \mathrm{NP}_3 \cdot \times(\mathrm{V}_2 \cdot \mathrm{NP}_4)))).$$

Note the use of embedded lock operators (the two rightmost occurrences). The rightmost and the leftmost occurrences of the lock operator implement our restriction against modifiers appearing in between the verbal and the object position. The occurrence of the lock operator in the middle of the IDL-expression prevents any of the modifiers $\mathtt{PP}_1$ and $\mathtt{PP}_2$ from modifying elements appearing within the subordinate clause. Observe that when we generalize the above examples by embedding $n$ subordinate clauses, the corresponding word lattice will grow exponentially in $n$, while the IDL-expression has linear size in $n$.

## 5. IDL-Graphs

Although IDL-expressions may be easily composed by linguists, they do not allow a direct algorithmic interpretation for efficient recognition of strings. We therefore define an equivalent but lower-level representation for IDL-expressions, which we call IDL-graphs. For this purpose, we exploit a specific kind of edge-labelled acyclic graphs with ranked nodes. We first introduce our notation, and then define the encoding function from IDL-expressions to IDL-graphs.

The graphs we use are denoted by tuples $(V, E, v_s, v_e, \lambda, r)$, where:

- $V$ and $E$ are finite sets of vertices and edges, respectively;
- $v_s$ and $v_e$ are special vertices in $V$ called the **start** and the **end** vertices, respectively;
- $\lambda$ is the edge-labelling function, mapping $E$ into the alphabet $\Sigma \cup \{\varepsilon, \vdash, \dashv\}$;
- $r$ is the vertex-ranking function, mapping $V$ to $\mathbb{N}$, the set of non-negative integer numbers.

Label $\varepsilon$ indicates that an edge does not consume any input symbols. Edge labels $\vdash$ and $\dashv$ have the same meaning, but they additionally encode that we are at the start or end, respectively, of what corresponds to an I operator. More precisely, let $\pi$ be an IDL-expression headed by an occurrence of the I operator and let $\gamma(\pi)$ be the associated IDL-graph. We use edges labelled by $\vdash$ to connect the start vertex of $\gamma(\pi)$ with the start vertices of all the subgraphs encoding the arguments of I. Similarly, we use edges labelled by $\dashv$ to connect all the end vertices of the subgraphs encoding the arguments of I with the end vertex of $\gamma(\pi)$. Edge labels $\vdash$ and $\dashv$ are needed in the next section to distinguish occurrences of the I operator from occurrences of the D and L operators. Finally, the function $r$ ranks each vertex according to how deeply it is embedded into (the encoding of) expressions headed by an occurrence of the L operator. As we will see later, this information is necessary for processing "locked" vertices with the correct priority.

We can now map an IDL-expression into the corresponding IDL-graph.

**Definition 3** *Let $\Sigma$ be some finite alphabet, and let $j$ be a non-negative integer number. Each IDL-expression $\pi$ over $\Sigma$ is associated with some graph $\gamma_j(\pi) = (V, E, v_s, v_e, \lambda, r)$ specified as follows:*

(i) *if $\pi = a$, $a \in \Sigma \cup \{\mathcal{E}\}$, let $v_s, v_e$ be new nodes; we have*

   (a) $V = \{v_s, v_e\}$,
   (b) $E = \{(v_s, v_e)\}$,
   (c) $\lambda((v_s, v_e)) = a$ *for* $a \in \Sigma$ *and* $\lambda((v_s, v_e)) = \varepsilon$ *for* $a = \mathcal{E}$,
   (d) $r(v_s) = r(v_e) = j$;

(ii) *if $\pi = \times(\pi')$ with $\gamma_{j+1}(\pi') = (V', E', v'_s, v'_e, \lambda', r')$, let $v_s, v_e$ be new nodes; we have*

   (a) $V = V' \cup \{v_s, v_e\}$,
   (b) $E = E' \cup \{(v_s, v'_s), (v'_e, v_e)\}$,
   (c) $\lambda(e) = \lambda'(e)$ *for* $e \in E'$, $\lambda((v_s, v'_s)) = \lambda((v'_e, v_e)) = \varepsilon$,
   (d) $r(v) = r'(v)$ *for* $v \in V'$, $r(v_s) = r(v_e) = j$;

(iii) *if $\pi = \vee(\pi_1, \pi_2, \ldots, \pi_n)$ with $\gamma_j(\pi_i) = (V_i, E_i, v_{i,s}, v_{i,e}, \lambda_i, r_i)$, $1 \leq i \leq n$, let $v_s, v_e$ be new nodes; we have*

   (a) $V = \cup_{i=1}^n V_i \cup \{v_s, v_e\}$,
   (b) $E = \cup_{i=1}^n E_i \cup \{(v_s, v_{i,s}) \mid 1 \leq i \leq n\} \cup \{(v_{i,e}, v_e) \mid 1 \leq i \leq n\}$,
   (c) $\lambda(e) = \lambda_i(e)$ *for* $e \in E_i$, $\lambda((v_s, v_{i,s})) = \lambda((v_{i,e}, v_e)) = \varepsilon$ *for* $1 \leq i \leq n$,
   (d) $r(v) = r_i(v)$ *for* $v \in V_i$, $r(v_s) = r(v_e) = j$;

(iv) *if $\pi = \|(\pi_1, \pi_2, \ldots, \pi_n)$ with $\gamma_j(\pi_i) = (V_i, E_i, v_{i,s}, v_{i,e}, \lambda_i, r_i)$, $1 \leq i \leq n$, let $v_s, v_e$ be new nodes; we have*

   (a) $V = \cup_{i=1}^n V_i \cup \{v_s, v_e\}$,
   (b) $E = \cup_{i=1}^n E_i \cup \{(v_s, v_{i,s}) \mid 1 \leq i \leq n\} \cup \{(v_{i,e}, v_e) \mid 1 \leq i \leq n\}$,

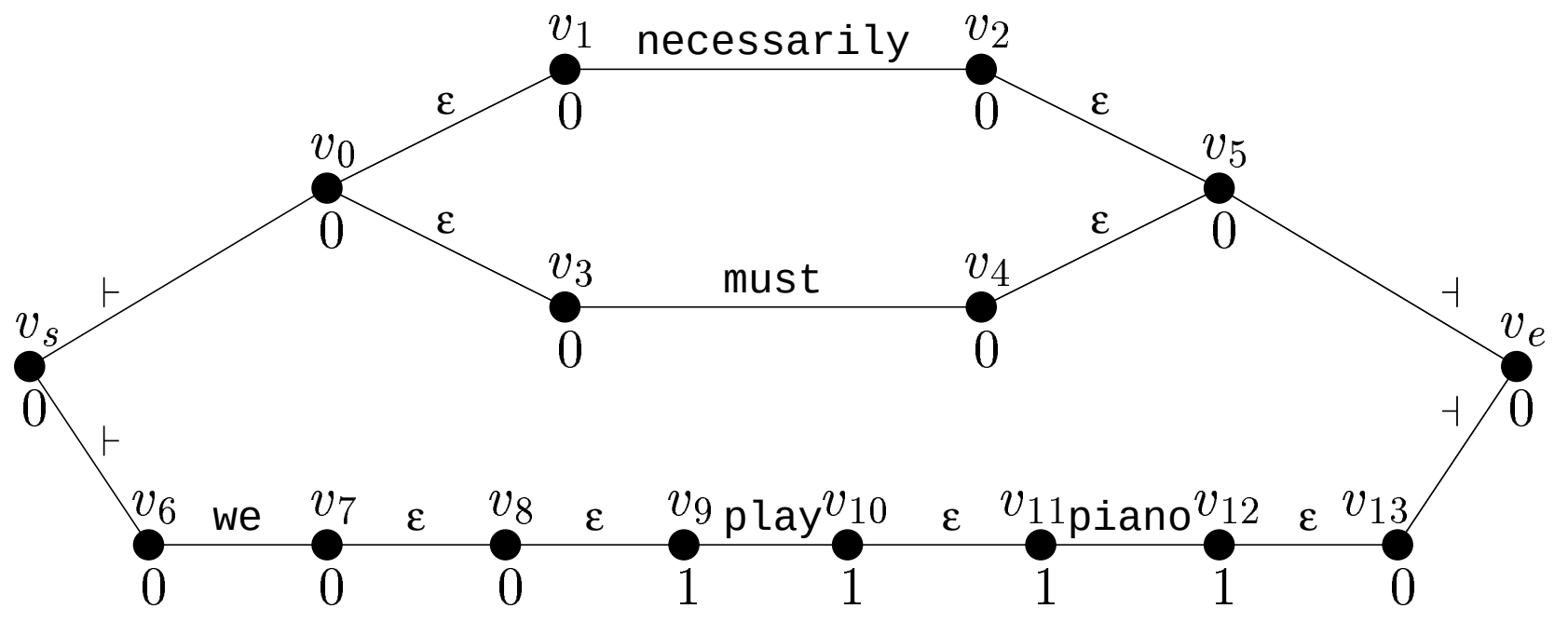

Figure 1: The IDL-graph associated with the IDL-expression $\|(\vee(\texttt{necessarily}, \texttt{must}), \texttt{we} \cdot \times(\texttt{play} \cdot \texttt{piano}))$.

(c) $\lambda(e) = \lambda_i(e)$ *for* $e \in E_i$, $\lambda((v_s, v_{i,s})) = \vdash$ *and* $\lambda((v_{i,e}, v_e)) = \dashv$ *for* $1 \leq i \leq n$,

(d) $r(v) = r_i(v)$ *for* $v \in V_i$, $r(v_s) = r(v_e) = j$;

(v) *if* $\pi = \pi_1 \cdot \pi_2$ *with* $\gamma_j(\pi_i) = (V_i, E_i, v_{i,s}, v_{i,e}, \lambda_i, r_i)$, $i \in \{1, 2\}$, *let* $v_s = v_{1,s}$ *and* $v_e = v_{2,e}$; *we have*

(a) $V = V_1 \cup V_2$,

(b) $E = E_1 \cup E_2 \cup \{(v_{1,e}, v_{2,s})\}$,

(c) $\lambda(e) = \lambda_i(e)$ *for* $e \in E_i$ *for* $i \in \{1, 2\}$, $\lambda((v_{1,e}, v_{2,s})) = \varepsilon$,

(d) $r(v) = r_i(v)$ *for* $v \in V_i$, $i \in \{1, 2\}$.

*We let* $\gamma(\pi) = \gamma_0(\pi)$. *An* **IDL-graph** *is a graph that has the form* $\gamma(\pi)$ *for some IDL-expression* $\pi$ *over* $\Sigma$.

Figure 1 presents the IDL-graph $\gamma(\pi)$, where $\pi$ is IDL-expression (2).

We now introduce the important notion of cut of an IDL-graph. This notion is needed to define the language described by an IDL-graph, so that we can talk about equivalence between IDL-expressions and IDL-graphs. At the same time, this notion will play a crucial role in the specification of our parsing algorithm for IDL-graphs in the next section. Let us fix some IDL-expression $\pi$ and let $\gamma(\pi) = (V, E, v_s, v_e, \lambda, r)$ be the associated IDL-graph. Intuitively speaking, a cut through $\gamma(\pi)$ is a set of vertices that we might reach simultaneously when traversing $\gamma(\pi)$ from the start vertex to the end vertex, following the different branches as prescribed by the encoded I, D and L operators, and in an attempt to produce a string of $L(\pi)$.

In what follows we view $V$ as a finite alphabet, and we define the set $\hat{V}$ to contain those strings over $V$ in which each symbol occurs at most once. Therefore $\hat{V}$ is a finite set and for each string $c \in \hat{V}$ we have $|c| \leq |V|$. If we assume that the outgoing edges of each vertex in an IDL-graph are linearly ordered, we can represent cuts in a canonical way by means of strings in $\hat{V}$ as defined below.

Let $r$ be the ranking function associated with $\gamma(\pi)$. We write $c[v_1 \cdots v_m]$ to denote a string $c \in \hat{V}$ satisfying the following conditions:

- $c$ has the form $xv_1 \cdots v_m y$ with $x, y \in \hat{V}$ and $v_i \in V$ for $1 \leq i \leq m$; *and*
- for each vertex $v$ within $c$ and for each $i$, $1 \leq i \leq m$, we have $r(v) \leq r(v_i)$.

In words, $c[v_1 \cdots v_m]$ indicates that vertices $v_1, \ldots, v_m$ occur adjacent in $c$ and have the maximal rank among all vertices within string $c$. Let $c[v_1 \cdots v_m] = xv_1 \cdots v_m y$ be a string defined as above and let $v'_1 \cdots v'_{m'} \in \hat{V}$ be a second string such that no symbol $v'_i$, $1 \leq i \leq m'$, appears in $x$ or $y$. We write $c[v_1 \cdots v_m := v'_1 \cdots v'_{m'}]$ to denote the string $xv'_1 \cdots v'_{m'} y \in \hat{V}$.

The reason we distinguish the vertices with maximal rank from those with lower rank is that the former correspond with subexpressions that are nested deeper within subexpressions headed by the L operator. As a substring originating within the scope of an occurrence of the lock operator cannot be interleaved with symbols originating outside that scope, we should terminate the processing of all vertices with higher rank before resuming processing of those with lower rank.

We now define a relation that plays a crucial role in the definition of the notion of cut, as well as in the specification of our parsing algorithm.

**Definition 4** *Let $\Sigma$ be some finite alphabet, let $\pi$ be an IDL-expression over $\Sigma$, and let $\gamma(\pi) = (V, E, v_s, v_e, \lambda, r)$ be its associated IDL-graph. The relation $\Delta_{\gamma(\pi)} \subseteq \hat{V} \times (\Sigma \cup \{\varepsilon\}) \times \hat{V}$ is the smallest satisfying all of the following conditions:*

(i) *for each $c[v] \in \hat{V}$ and $(v, v') \in E$ with $\lambda((v, v')) = X \in \Sigma \cup \{\varepsilon\}$, we have*

$$(c[v], X, c[v := v']) \quad \in \quad \Delta_{\gamma(\pi)}; \tag{4}$$

(ii) *for each $c[v] \in \hat{V}$ with the outgoing edges of $v$ being exactly $(v, v_1), \ldots, (v, v_n) \in E$, in this order, and with $\lambda((v, v_i)) = \vdash$, $1 \leq i \leq n$, we have*

$$(c[v], \varepsilon, c[v := v_1 \cdots v_n]) \quad \in \quad \Delta_{\gamma(\pi)}; \tag{5}$$

(iii) *for each $c[v_1 \cdots v_n] \in \hat{V}$ with the incoming edges of some $v \in V$ being exactly $(v_1, v), \ldots, (v_n, v) \in E$, in this order, and with $\lambda((v_i, v)) = \dashv$, $1 \leq i \leq n$, we have*

$$(c[v_1 \cdots v_n], \varepsilon, c[v_1 \cdots v_n := v]) \quad \in \quad \Delta_{\gamma(\pi)}. \tag{6}$$

Henceforth, we will abuse notation by writing $\Delta_\pi$ in place of $\Delta_{\gamma(\pi)}$. Intuitively speaking, relation $\Delta_\pi$ will be used to simulate a one-step move over IDL-graph $\gamma(\pi)$. Condition (4) refers to moves that follow a single edge in the graph, labelled by a symbol from the alphabet or by the empty string. This move is exploited, e.g., upon visiting a vertex at the start of a subgraph that encodes an IDL-expression headed by an occurrence of the D operator. In this case, each outgoing edge represents a possible next move, but at most one edge can be chosen. Condition (5) refers to moves that simultaneously follow all edges emanating from the vertex at hand. This is used when processing a vertex at the start of a subgraph that encodes an IDL-expression headed by an occurrence of the I operator. In fact, in accordance with the given semantics, all possible argument expressions must be evaluated in parallel by a single computation. Finally, Condition (6) refers to a move that can be read as the complement of the previous type of move.

Examples of elements in $\Delta_\pi$ in the case of Figure 1 are $(v_s, \varepsilon, v_0v_6)$ following Condition (5) and $(v_5v_{13}, \varepsilon, v_e)$ following Condition (6), which start and end the evaluation of the occurrence of the I operator. Other elements are $(v_0v_6, \varepsilon, v_1v_6)$, $(v_1v_9, \texttt{play}, v_1v_{10})$ and $(v_1v_{13}, \texttt{necessarily}, v_2v_{13})$ following Condition (4). Note that, e.g., $(v_1v_{10}, \texttt{necessarily}, v_2v_{10})$ is *not* an element of $\Delta_\pi$, as $v_9$ has higher rank than $v_1$.

We are now ready to define the notion of cut.

**Definition 5** *Let $\Sigma$ be some finite alphabet, let $\pi$ be an IDL-expression over $\Sigma$, and let $\gamma(\pi) = (V, E, v_s, v_e, \lambda, r)$ be its associated IDL-graph. The set of all* **cuts** *of $\gamma(\pi)$, written $\mathsf{cut}(\gamma(\pi))$, is the smallest subset of $\hat{V}$ satisfying the following conditions:*

(i) *string $v_s$ belongs to $\mathsf{cut}(\gamma(\pi))$;*

(ii) *for each $c \in \mathsf{cut}(\gamma(\pi))$ and $(c, X, c') \in \Delta_\pi$, string $c'$ belongs to $\mathsf{cut}(\gamma(\pi))$.*

Henceforth, we will abuse notation by writing $\mathsf{cut}(\pi)$ for $\mathsf{cut}(\gamma(\pi))$. As already remarked, we can interpret a cut $v_1v_2 \cdots v_k \in \mathsf{cut}(\pi)$, $v_i \in V$ for $1 \leq i \leq k$, as follows. In the attempt to generate a string in $L(\pi)$, we traverse several paths of the IDL-graph $\gamma(\pi)$. This corresponds to the "parallel" evaluation of some of the subexpressions of $\pi$, and each $v_i$ in $v_1v_2 \cdots v_k$ refers to one such subexpression. Thus, $k$ provides the number of evaluations that we are carrying out in parallel at the point of the computation represented by the cut. Note however that, when drawing a straight line across a planar representation of an IDL-graph, separating the start vertex from the end vertex, the set of vertices that we can identify is not necessarily a cut.[4] In fact, as we have already explained when discussing relation $\Delta_\pi$, only one path is followed at the start of a subgraph that encodes an IDL-expression headed by an occurrence of the D operator. Furthermore, even if several arcs are to be followed at the start of a subgraph that encodes an IDL-expression headed by an occurrence of the I operator, some combinations of vertices will not satisfy the definition of cut when there are L operators within those argument expressions. These observations will be more precisely addressed in Section 7, where we will provide a mathematical analysis of the complexity of our algorithm.

Examples of cuts in the case of Figure 1 are $v_s$, $v_e$, $v_0v_6$, $v_1v_6$, $v_3v_6$, $v_0v_7$, etc. Strings such as $v_1v_3$ are *not* cuts, as $v_1$ and $v_3$ belong to two disjoint subgraphs with sets of vertices $\{v_1, v_2\}$ and $\{v_3, v_4\}$, respectively, each of which corresponds to a different argument of an occurrence of the disjunction operator.

Given the notion of cut, we can associate a finite language with each IDL-graph and talk about equivalence with IDL-expressions. Let $\pi$ be an IDL-expression over $\Sigma$, and let $\gamma(\pi) = (V, E, v_s, v_e, \lambda, r)$ be the associated IDL-graph. Let also $c, c' \in \mathsf{cut}(\pi)$ and $w \in \Sigma^*$. We write $w \in L(c, c')$ if there exists $q \geq |w|$, $X_i \in \Sigma \cup \{\varepsilon\}$, $1 \leq i \leq q$, and $c_i \in \mathsf{cut}(\pi)$, $0 \leq i \leq q$, such that $X_1 \cdots X_q = w$, $c_0 = c$, $c_q = c'$ and $(c_{i-1}, X_i, c_i) \in \Delta_\pi$ for $1 \leq i \leq q$.

4. The pictorial representation mentioned above comes close to a different definition of cut that is standard in the literature on graph theory and operating research. The reader should be aware that this standard graph-theoretic notion of "cut" is different from the one introduced in this paper.

We also assume that $L(c, c) = \{\varepsilon\}$. We can then show that $L(v_s, v_e) = L(\pi)$, i.e., the language generated by the IDL-expression $\pi$ is the same as the language that we obtain in a traversal of the IDL-graph $\gamma(\pi)$, as described above, starting from cut $v_s$ and ending in cut $v_e$. The proof of this property is rather long and does not add much to the already provided intuition underlying the definitions in this section; therefore we will omit it.

We close this section with an informal discussion of relation $\Delta_\pi$ and the associated notion of cut. Observe that Definition 4 and Definition 5 implicitly define a nondeterministic finite automaton. Again, we refer the reader to Harrison (1978) for a definition of finite automata. The states of the automaton are the cuts in $\mathsf{cut}(\pi)$ and its transitions are given by the elements of $\Delta_\pi$. The initial state of the automaton is the cut $v_s$, and the final state is the cut $v_e$. It is not difficult to see that from every state of the automaton one can always reach the final state. Furthermore, the language recognized by such an automaton is precisely the language $L(v_s, v_e)$ defined above. However, we remark here that such an automaton will never be constructed by our parsing algorithm, as emphasized in the next section.

## 6. CFG Parsing of IDL-Graphs

We start this section with a brief overview of the Earley algorithm (Earley, 1970), a well-known tabular method for parsing input strings according to a given CFG. We then reformulate the Earley algorithm in order to parse IDL-graphs. As already mentioned in the introduction, while parsing is traditionally defined for input consisting of a single string, we here conceive parsing as a process that can be carried out on an input device representing a language, i.e., a set of strings.

Let $G = (N, \Sigma, P, S)$ be a CFG, and let $w = a_1 \cdots a_n \in \Sigma^*$ be an input string to be parsed. Standard implementations of the Earley algorithm (Graham & Harrison, 1976) use so called parsing items to record partial results of the parsing process on $w$. A parsing item has the form $[A \to \alpha \bullet \beta, i, j]$, where $A \to \alpha\beta$ is a production of $G$ and $i$ and $j$ are indices identifying a substring $a_{i+1} \cdots a_j$ of $w$. Such a parsing item is constructed by the algorithm if and only if there exist a string $\gamma \in (N \cup \Sigma)^*$ and two derivations in $G$ having the form

$$\begin{array}{lll} S & \Rightarrow_G^* & a_1 \cdots a_i A \gamma \\ & \Rightarrow_G & a_1 \cdots a_i \alpha\beta\gamma; \\ \alpha & \Rightarrow_G^* & a_{i+1} \cdots a_j. \end{array}$$

The algorithm accepts $w$ if and only if it can construct an item of the form $[S \to \alpha \bullet, 0, n]$, for some production $S \to \alpha$ of $G$. Figure 2 provides an abstract specification of the algorithm expressed as a deduction system, following Shieber, Schabes, and Pereira (1995). Inference rules specify the types of steps that the algorithm can apply in constructing new items.

Rule (7) in Figure 2 serves as an initialization step, constructing all items that can start analyses for productions with the start symbol $S$ in the right-hand side. Rule (8) is very similar in purpose: it constructs all items that can start analyses for productions with nonterminal $B$ in the left-hand side, provided that $B$ is the next nonterminal in some existing item for which an analysis is to be found. Rule (9) matches a terminal $a$ in an item with an input symbol, and the new item signifies that a larger part of the right-hand side has been matched to a larger part of the input. Finally, Rule (10) combines two partial

$$\frac{}{[S \to \bullet\, \alpha, 0, 0]} \left\{ \; S \to \alpha \right. \tag{7}$$

$$\frac{[A \to \alpha \bullet B\beta, i, j]}{[B \to \bullet\, \gamma, j, j]} \left\{ \; B \to \gamma \right. \tag{8}$$

$$\frac{[A \to \alpha \bullet a\beta, i, j]}{[A \to \alpha a \bullet \beta, i, j+1]} \left\{ \; a = a_{j+1} \right. \tag{9}$$

$$\frac{\begin{array}{l}[A \to \alpha \bullet B\beta, i, j] \\ [B \to \gamma \,\bullet, j, k]\end{array}}{[A \to \alpha B \bullet \beta, i, k]} \tag{10}$$

Figure 2: Abstract specification of the parsing algorithm of Earley for an input string $a_1 \cdots a_n$. The algorithm accepts $w$ if and only if it can construct an item of the form $[S \to \alpha\,\bullet, 0, n]$, for some production $S \to \alpha$ of $G$.

analyses, the second of which represents an analysis for symbol $B$, by which the analysis represented by the first item can be extended.

We can now move to our algorithm for IDL-graph parsing using a CFG. The algorithm makes use of relation $\Delta_\pi$ from Definition 4, but this does not mean that the relation is fully computed before invoking the algorithm. We instead compute elements of $\Delta_\pi$ "on-the-fly" when we first visit a cut, and cache these elements for possible later use. This has the advantage that, when parsing an input IDL-graph, our algorithm processes only those portions of the graph that represent prefixes of strings that are generated by the CFG at hand. In practical cases, the input IDL-graph is never completely unfolded, so that the compactness of the proposed representation is preserved to a large extent.

An alternative way of viewing our algorithm is this. We have already informally discussed in Section 5 how relation $\Delta_\pi$ implicitly defines a nondeterministic finite automaton whose states are the elements of $\mathsf{cut}(\pi)$ and whose transitions are the elements of $\Delta_\pi$. We have also mentioned that such an automaton precisely recognizes the finite language $L(\pi)$. From this perspective, our algorithm can be seen as a standard lattice parsing algorithm, discussed in Section 2. What must be emphasized here is that we do not precompute the above finite automaton prior to parsing. Our approach consists in a lazy evaluation of the transitions of the automaton, on the basis of a demand on the part of the parsing process. In contrast with our approach, full expansion of the finite automaton before parsing has several disadvantages. Firstly, although a finite automaton generating a finite language might be considerably smaller than a representation of the language itself consisting of a list of all its elements, it is easy to see that there are cases in which the finite automaton might have size exponentially larger than the corresponding IDL-expression (see also the discussion in Section 2). In such cases, full expansion destroys the compactness of IDL-expressions, which is the main motivation for the use of our formalism in hybrid surface generation systems, as discussed in the introduction. Furthermore, full expansion of the automaton is also computationally unattractive, since it may lead to unfolding of parts of the input IDL-graph that will never be processed by the parsing algorithm.

Let $G = (N, \Sigma, P, S)$ be a CFG and let $\pi$ be some input IDL-expression. The algorithm uses parsing items of the form $[A \to \alpha \bullet \beta, c_1, c_2]$, with $A \to \alpha\beta$ a production in $P$ and $c_1, c_2 \in \mathsf{cut}(\pi)$. These items have the same meaning as those used in the original Earley algorithm, but now they refer to strings in the languages $L(v_s, c_1)$ and $L(c_1, c_2)$, where $v_s$ is the start vertex of IDL-graph $\gamma(\pi)$. (Recall from Section 5 that $L(c, c')$, $c, c' \in \mathsf{cut}(\pi)$, is the set of strings whose symbols can be consumed in any traversal of $\gamma(\pi)$ starting from cut $c$ and ending in cut $c'$.) We also use items of the forms $[c_1, c_2]$ and $[a, c_1, c_2]$, $a \in \Sigma$, $c_1, c_2 \in \mathsf{cut}(\pi)$. This is done in order to by-pass traversals of $\gamma(\pi)$ involving a sequence of zero or more triples of the form $(c_1, \varepsilon, c_2) \in \Delta_\pi$, followed by a triple of the form $(c_1, a, c_2) \in \Delta_\pi$. Figure 3 presents an abstract specification of the algorithm, again using a set of inference rules. The issues of control flow and implementation are deferred to the next section.

In what follows, let $v_s$ and $v_e$ be the start and end vertices of IDL-graph $\gamma(\pi)$, respectively. Rules (11), (12) and (15) in Figure 3 closely resemble Rules (7), (8) and (10) of the original Earley algorithm, as reported in Figure 2. Rules (13), (16) and (17) have been introduced for the purpose of efficiently computing traversals of $\gamma(\pi)$ involving a sequence of zero or more triples of the form $(c_1, \varepsilon, c_2) \in \Delta_\pi$, followed by a triple of the form $(c_1, a, c_2) \in \Delta_\pi$, as already mentioned. Once one such traversal has been computed, the fact is recorded through some item of the form $[a, c_1, c_2]$, avoiding later recomputation. Rule (14) closely resembles Rule (9) of the original Earley algorithm. Finally, by computing traversals of $\gamma(\pi)$ involving triples of the form $(c_1, \varepsilon, c_2) \in \Delta_\pi$ only, Rule (18) may derive items of the form $[S \to \alpha \bullet, v_s, v_e]$; the algorithm accepts the input IDL-graph if and only if any such item can be derived by the inference rules.

We now turn to the discussion of the correctness of the algorithm in Figure 3. Our algorithm derives a parsing item $[A \to \alpha \bullet \beta, c_1, c_2]$ if and only if there exist a string $\gamma \in (N \cup \Sigma)^*$, integers $i, j$ with $0 \le i \le j$, and $a_1 a_2 \cdots a_j \in \Sigma^*$ such that the following conditions are all satisfied:

- $a_1 \cdots a_i \in L(v_s, c_1)$;
- $a_{i+1} \cdots a_j \in L(c_1, c_2)$; *and*
- there exist two derivations in $G$ of the form

$$\begin{aligned} S &\Rightarrow_G^* a_1 \cdots a_i A \gamma \\ &\Rightarrow_G a_1 \cdots a_i \alpha\beta\gamma \\ \alpha &\Rightarrow_G^* a_{i+1} \cdots a_j. \end{aligned}$$

The above statement closely resembles the existential condition previously discussed for the original Earley algorithm, and can be proved using arguments similar to those presented for

$$\frac{}{[S \to \bullet\ \alpha, v_s, v_s]} \left\{\ S \to \alpha \right. \tag{11}$$

$$\frac{[A \to \alpha \bullet B\beta, c_1, c_2]}{[B \to \bullet\ \gamma, c_2, c_2]} \left\{\ B \to \gamma \right. \tag{12}$$

$$\frac{[A \to \alpha \bullet a\beta, c_1, c_2]}{[c_2, c_2]} \tag{13}$$

$$\frac{\begin{array}{l}[A \to \alpha \bullet a\beta, c_1, c_2] \\ [a, c_2, c_3]\end{array}}{[A \to \alpha a \bullet \beta, c_1, c_3]} \tag{14}$$

$$\frac{\begin{array}{l}[A \to \alpha \bullet B\beta, c_1, c_2] \\ [B \to \gamma \bullet, c_2, c_3]\end{array}}{[A \to \alpha B \bullet \beta, c_1, c_3]} \tag{15}$$

$$\frac{[c_1, c_2]}{[c_1, c_3]} \left\{\ (c_2, \epsilon, c_3) \in \Delta_\pi \right. \tag{16}$$

$$\frac{[c_1, c_2]}{[a, c_1, c_3]} \left\{\begin{array}{l} (c_2, a, c_3) \in \Delta_\pi, \\ a \in \Sigma \end{array}\right. \tag{17}$$

$$\frac{[S \to \alpha \bullet, c_0, c_1]}{[S \to \alpha \bullet, c_0, c_2]} \left\{\ (c_1, \epsilon, c_2) \in \Delta_\pi \right. \tag{18}$$

Figure 3: An abstract specification of the parsing algorithm for IDL-graphs. The algorithm accepts the IDL-graph $\gamma(\pi)$ if and only if some item having the form $[S \to \alpha \bullet, v_s, v_e]$ can be derived by the inference rules, where $S \to \alpha$ is a production of $G$ and $v_s$ and $v_e$ are the start and end vertices of $\gamma(\pi)$, respectively.

instance by Aho and Ullman (1972) and by Graham and Harrison (1976); we will therefore omit a complete proof here. Note that the correctness of the algorithm in Figure 3 directly follows from the above statement, by taking item $[A \rightarrow \alpha \bullet \beta, c_1, c_2]$ to be of the form $[S \rightarrow \alpha \bullet, v_s, v_e]$ for some production $S \rightarrow \alpha$ from $G$.

## 7. Complexity and Implementation

In this section we provide a computational analysis of our parsing algorithm for IDL-graphs. The analysis is based on the development of a tight upper bound on the number of possible cuts admitted by an IDL-graph. We also discuss two possible implementations for the parsing algorithm.

We need to introduce some notation. Let $\pi$ be an IDL-expression and let $\gamma(\pi) = (V, E, v_s, v_e, \lambda, r)$ be the associated IDL-graph. A vertex $v \in V$ is called L-free in $\gamma(\pi)$ if, for every subexpression $\pi'$ of $\pi$ such that $\gamma_j(\pi') = (V', E', v'_s, v'_e, \lambda', r')$ for some $j$, $V' \subseteq V$, $E' \subseteq E$, and such that $v \in V'$, we have that $\pi'$ is not of the form $\times(\pi'')$. In words, a vertex is L-free in $\gamma(\pi)$ if it does not belong to a subgraph of $\gamma(\pi)$ that encodes an IDL-expression headed by an L operator. When $\gamma(\pi)$ is understood from the context, we write L-free in place of L-free in $\gamma(\pi)$. We write $\mathsf{0\text{-}cut}(\pi)$ to denote the set of all cuts in $\mathsf{cut}(\pi)$ that only contain vertices that are L-free in $\gamma(\pi)$. We now introduce two functions that will be used later in the complexity analysis of our algorithm. For a cut $c \in \mathsf{cut}(\pi)$ we write $|c|$ to denote the length of $c$, i.e., the number of vertices in the cut.

**Definition 6** *Let $\pi$ be an IDL-expression. Functions* $\mathsf{width}$ *and* $\mathsf{0\text{-}width}$ *are specified as follows:*

$$\begin{aligned}\mathsf{width}(\pi) &= \max_{c \in \mathsf{cut}(\pi)} |c|, \\ \mathsf{0\text{-}width}(\pi) &= \max_{c \in \mathsf{0\text{-}cut}(\pi)} |c|.\end{aligned}$$

Function $\mathsf{width}$ provides the maximum length of a cut through $\gamma(\pi)$. This quantity gives the maximum number of subexpressions of $\pi$ that need to be evaluated in parallel when generating a string in $L(\pi)$. Similarly, function $\mathsf{0\text{-}width}$ provides the maximum length of a cut through $\gamma(\pi)$ that only includes L-free nodes.

Despite the fact that $\mathsf{cut}(\pi)$ is always a finite set, a computation of functions $\mathsf{width}$ and $\mathsf{0\text{-}width}$ through a direct computation of $\mathsf{cut}(\pi)$ and $\mathsf{0\text{-}cut}(\pi)$ is not practical, since these sets may have exponential size in the number of vertices of $\gamma(\pi)$. The next characterization provides a more efficient way to compute the above functions, and will be used in the proof of Lemma 2 below.

**Lemma 1** *Let $\pi$ be an IDL-expression. The quantities* $\mathsf{width}(\pi)$ *and* $\mathsf{0\text{-}width}(\pi)$ *satisfy the following equations:*

(i) *if $\pi = a$, $a \in \Sigma \cup \{\mathcal{E}\}$, we have*

$$\begin{aligned}\mathsf{width}(\pi) &= 1, \\ \mathsf{0\text{-}width}(\pi) &= 1;\end{aligned}$$

(ii) *if* $\pi = \times(\pi')$ *we have*

$$\begin{aligned}\mathsf{width}(\pi) &= \mathsf{width}(\pi'),\\ \mathsf{0\text{-}width}(\pi) &= 1;\end{aligned}$$

(iii) *if* $\pi = \vee(\pi_1, \pi_2, \ldots, \pi_n)$ *we have*

$$\begin{aligned}\mathsf{width}(\pi) &= \max_{i=1}^{n} \mathsf{width}(\pi_i),\\ \mathsf{0\text{-}width}(\pi) &= \max_{i=1}^{n} \mathsf{0\text{-}width}(\pi_i);\end{aligned}$$

(iv) *if* $\pi = \|(\pi_1, \pi_2, \ldots, \pi_n)$ *we have*

$$\begin{aligned}\mathsf{width}(\pi) &= \max_{j=1}^{n} (\mathsf{width}(\pi_j) + \sum_{i:1\leq i\leq n \wedge i\neq j} \mathsf{0\text{-}width}(\pi_i)),\\ \mathsf{0\text{-}width}(\pi) &= \sum_{j=1}^{n} \mathsf{0\text{-}width}(\pi_j);\end{aligned}$$

(v) *if* $\pi = \pi_1 \cdot \pi_2$ *we have*

$$\begin{aligned}\mathsf{width}(\pi) &= \max\{\mathsf{width}(\pi_1), \mathsf{width}(\pi_2)\},\\ \mathsf{0\text{-}width}(\pi) &= \max\{\mathsf{0\text{-}width}(\pi_1), \mathsf{0\text{-}width}(\pi_2)\}.\end{aligned}$$

*Proof.* All of the equations in the statement of the lemma straightforwardly follow from the definitions of $\Delta_\pi$ and $\mathsf{cut}(\pi)$ (Definitions 4 and 5, respectively). Here we develop at length only two cases and leave the remainder of the proof to the reader. In what follows we assume that $\gamma(\pi) = (V, E, v_s, v_e, \lambda, r)$.

In case $\pi = \vee(\pi_1, \pi_2, \ldots, \pi_n)$, let $\gamma(\pi_i) = (V_i, E_i, v_{i,s}, v_{i,e}, \lambda_i, r_i)$, $1 \leq i \leq n$. From Definition 4 we have $(v_s, \varepsilon, v_{i,s}) \in \Delta_\pi$ and $(v_{i,e}, \varepsilon, v_e) \in \Delta_\pi$, for every $i$, $1 \leq i \leq n$. Thus we have $\mathsf{cut}(\pi) = \cup_{i=1}^{n}\mathsf{cut}(\pi_i) \cup \{v_s, v_e\}$ and, since both $v_s$ and $v_e$ are L-free in $\gamma(\pi)$, $\mathsf{0\text{-}cut}(\pi) = \cup_{i=1}^{n}\mathsf{0\text{-}cut}(\pi_i) \cup \{v_s, v_e\}$. This provides the relations in (iii).

In case $\pi = \|(\pi_1, \pi_2, \ldots, \pi_n)$, let $\gamma(\pi_i) = (V_i, E_i, v_{i,s}, v_{i,e}, \lambda_i, r_i)$, $1 \leq i \leq n$. From Definition 4 we have $(v_s, \varepsilon, v_{1,s} \cdots v_{n,s}) \in \Delta_\pi$ and $(v_{1,e} \cdots v_{n,e}, \varepsilon, v_e) \in \Delta_\pi$. Thus every $c \in \mathsf{cut}(\pi)$ must belong to $\{v_s, v_e\}$ or must have the form $c = c_1 \cdots c_n$ with $c_i \in \mathsf{cut}(\pi_i)$ for $1 \leq i \leq n$. Since both $v_s$ and $v_e$ are L-free in $\gamma(\pi)$, we immediately derive

$$\mathsf{0\text{-}cut}(\pi) = \{v_s, v_e\} \cup \mathsf{0\text{-}cut}(\pi_1) \cdots \mathsf{0\text{-}cut}(\pi_n),$$

and hence $\mathsf{0\text{-}width}(\pi) = \sum_{j=1}^{n} \mathsf{0\text{-}width}(\pi_j)$. Now observe that, for each $c = c_1 \cdots c_n$ specified as above there can never be indices $i$ and $j$, $1 \leq i, j \leq n$ and $i \neq j$, and vertices $v_1$ and $v_2$ occurring in $c_i$ and $c_j$, respectively, such that neither $v_1$ nor $v_2$ are L-free in $\gamma(\pi)$.

We thereby derive

$$\begin{aligned}\mathsf{cut}(\pi) = \; & \{v_s, v_e\} \cup\\ & \mathsf{cut}(\pi_1)\mathsf{0\text{-}cut}(\pi_2) \cdots \mathsf{0\text{-}cut}(\pi_n) \cup\\ & \mathsf{0\text{-}cut}(\pi_1)\mathsf{cut}(\pi_2) \cdots \mathsf{0\text{-}cut}(\pi_n) \cup\\ & \vdots\\ & \mathsf{0\text{-}cut}(\pi_1)\mathsf{0\text{-}cut}(\pi_2) \cdots \mathsf{cut}(\pi_n).\end{aligned}$$

Hence we can write $\mathsf{width}(\pi) = \max_{j=1}^{n} (\mathsf{width}(\pi_j) + \sum_{i:1 \le i \le n \wedge i \ne j} \mathsf{0\text{-}width}(\pi_i))$. □

Now consider quantity $|\mathsf{cut}(\pi)|$, i.e., the number of different cuts in IDL-graph $\gamma(\pi)$. This quantity is obviously bounded from above by $|V|^{\mathsf{width}(\pi)}$. We now derive a tighter upper bound on this quantity.

**Lemma 2** *Let $\Sigma$ be a finite alphabet, let $\pi$ be an IDL-expression over $\Sigma$, and let $\gamma(\pi) = (V, E, v_s, v_e, \lambda, r)$ be its associated IDL-graph. Let also $k = \mathsf{width}(\pi)$. We have*

$$|\mathsf{cut}(\pi)| \quad \le \quad \left(\frac{|V|}{k}\right)^k .$$

*Proof.* We use below the following inequality. For any integer $h \ge 2$ and real values $x_i > 0$, $1 \le i \le h$, we have

$$\prod_{i=1}^{h} x_i \quad \le \quad \left(\frac{\sum_{i=1}^{h} x_i}{h}\right)^h . \tag{19}$$

In words, (19) states that the geometric mean is never larger than the arithmetic mean.

We prove (19) in the following equivalent form. For any real values $c > 0$ and $y_i$, $1 \le i \le h$ and $h \ge 2$, with $y_i > -c$ and $\sum_{i=1}^{h} y_i = 0$, we have

$$\prod_{i=1}^{h} (c + y_i) \quad \le \quad c^h . \tag{20}$$

We start by observing that if the $y_i$ are all equal to zero, then we are done. Otherwise there must be $i$ and $j$ with $1 \le i, j \le h$ such that $y_i y_j < 0$. Without loss of generality, we assume $i = 1$ and $j = 2$. Since $y_i y_j < 0$, we have

$$(c + y_1)(c + y_2) = c(c + y_1 + y_2) + y_1 y_2 < c(c + y_1 + y_2). \tag{21}$$

Since $\prod_{i=3}^{h} (c + y_i) > 0$, we have

$$(c + y_1)(c + y_2) \prod_{i=3}^{h} (c + y_i) \quad < \quad c(c + y_1 + y_2) \prod_{i=3}^{h} (c + y_i). \tag{22}$$

We now observe that the right-hand side of (22) has the same form as the left-hand side of (20), but with fewer $y_i$ that are non-zero. We can therefore iterate the above procedure, until all $y_i$ become zero valued. This concludes the proof of (19).

Let us turn to the proof of the statement of the lemma. Recall that each cut $c \in \mathsf{cut}(\pi)$ is a string over $V$ such that no vertex in $V$ has more than one occurrence in $c$, and $c$ is canonically represented, i.e., no other permutation of the vertices in $c$ is a possible cut. We will later prove the following claim.

*Claim.* Let $\pi$, $V$ and $k$ be as in the statement of the lemma. We can partition $V$ into subsets $V[\pi, j]$, $1 \le j \le k$, having the following property. For every $V[\pi, j]$, $1 \le j \le k$, and every pair of distinct vertices $v_1, v_2 \in V[\pi, j]$, $v_1$ and $v_2$ do not occur together in any cut $c \in \mathsf{cut}(\pi)$.

We can then write

$$
\begin{array}{rcll}
|\mathsf{cut}(\pi)| & \leq & \prod_{j=1}^{k} |V[\pi, j]| & \text{(by our claim and the canonical} \\
 & & & \text{representation of cuts)} \\
 & \leq & \left( \frac{\sum_{j=1}^{k} |V[\pi,j]|}{k} \right)^{k} & \text{(by (19))} \\
 & = & \left( \frac{|V|}{k} \right)^{k}. &
\end{array}
$$

To complete the proof of the lemma we now need to prove our claim above. We prove the following statement, which is a slightly stronger version of the claim. We can partition set $V$ into subsets $V[\pi, j]$, $1 \leq j \leq k = \mathsf{width}(\pi)$, having the following two properties:

- for every $V[\pi, j]$, $1 \leq j \leq k$, and every pair of distinct vertices $v_1, v_2 \in V[\pi, j]$, $v_1$ and $v_2$ do not occur together in any cut $c \in \mathsf{cut}(\pi)$;
- all vertices in $V$ that are L-free in $\gamma(\pi)$ are included in some $V[\pi, j]$, $1 \leq j \leq \mathsf{0\text{-}width}(\pi)$. (In other words, the sets $V[\pi, j]$, $\mathsf{0\text{-}width}(\pi) < j \leq \mathsf{width}(\pi)$, can only contain vertices that are not L-free in $\gamma(\pi)$.)

In what follows we use induction on $\#_{op}(\pi)$, the number of operator occurrences (I, D, L and concatenation) appearing within $\pi$.

*Base:* $\#_{op}(\pi) = 0$. We have $\pi = a$, with $a \in \Sigma \cup \{\mathcal{E}\}$, and $V = \{v_s, v_f\}$. Since $\mathsf{width}(\pi) = 1$, we set $V[\pi, 1] = V$. This satisfies our claim, since $\mathsf{cut}(\pi) = \{v_s, v_f\}$, all vertices in $V$ are L-free in $\gamma(\pi)$ and we have $\mathsf{0\text{-}width}(\pi) = 1$.

*Induction:* $\#_{op}(\pi) > 0$. We distinguish among three possible cases.

Case 1: $\pi = \vee(\pi_1, \pi_2, \ldots, \pi_n)$. Let $\gamma(\pi_i) = (V_i, E_i, v_{i,s}, v_{i,e}, \lambda_i, r_i)$, $1 \leq i \leq n$. By Lemma 1 we have $\mathsf{width}(\pi) = \max_{i=1}^{n} \mathsf{width}(\pi_i)$. For each $i$, $1 \leq i \leq n$, let us define $V[\pi_i, j] = \emptyset$ for every $j$ such that $\mathsf{width}(\pi_i) < j \leq \mathsf{width}(\pi)$. We can then set

$$
\begin{array}{rcl}
V[\pi, 1] & = & (\cup_{i=1}^{n} V[\pi_i, 1]) \cup \{v_s, v_e\}; \\
V[\pi, j] & = & \cup_{i=1}^{n} V[\pi_i, j], \quad \text{for } 2 \leq j \leq \mathsf{width}(\pi).
\end{array}
$$

The sets $V[\pi, j]$ define a partition of $V$, since $V = (\cup_{i=1}^{n} V_i) \cup \{v_s, v_e\}$ and, for each $i$, the sets $V[\pi_i, j]$ define a partition of $V_i$ by the inductive hypothesis. We now show that such a partition satisfies the two conditions in our statement.

Let $v_1$ and $v_2$ be two distinct vertices in some $V[\pi, j]$. We have already established in the proof of Lemma 1 that $\mathsf{cut}(\pi) = (\cup_{i=1}^{n} \mathsf{cut}(\pi_i)) \cup \{v_s, v_e\}$. If either $v_1$ or $v_2$ belongs to the set $\{v_s, v_e\}$, then $v_1$ and $v_2$ cannot occur in the same cut in $\mathsf{cut}(\pi)$, since the only cuts in $\mathsf{cut}(\pi)$ with vertices in the set $\{v_s, v_e\}$ are $v_s$ and $v_e$. Let us now consider the case $v_1, v_2 \in \cup_{i=1}^{n} V_i$. We can distinguish two subcases. In the first subcase, there exists $i$ such that $v_1, v_2 \in V[\pi_i, j]$. The inductive hypothesis states that $v_1$ and $v_2$ cannot occur in the same cut in $\mathsf{cut}(\pi_i)$, and hence cannot occur in the same cut in $\mathsf{cut}(\pi)$. In the second subcase, $v_1 \in V[\pi_i, j]$ and $v_2 \in V[\pi_{i'}, j]$ for distinct $i$ and $i'$. Then $v_1$ and $v_2$ must belong to different graphs $\gamma(\pi_i)$ and $\gamma(\pi_{i'})$, and hence cannot occur in the same cut in $\mathsf{cut}(\pi)$.

Furthermore, every vertex in $\cup_{i=1}^{n} V_i$ that is L-free in some $\gamma(\pi_i)$ belongs to some $V[\pi_i, j]$ with $1 \leq j \leq \mathsf{0\text{-}width}(\pi_i)$, by the inductive hypothesis. Since $\mathsf{0\text{-}width}(\pi) =$ $\max_{i=1}^{n}$ $\mathsf{0\text{-}width}(\pi_i)$ (Lemma 1) we can state that all vertices in $V$ that are L-free in $\gamma(\pi)$ belong to some $V[\pi, j]$, $1 \leq j \leq \mathsf{0\text{-}width}(\pi)$.
Case 2: $\pi = \times(\pi')$ or $\pi = \pi_1 \cdot \pi_2$. The proof is almost identical to that of Case 1, with $n = 1$ or $n = 2$, respectively.
Case 3: $\pi = \|(\pi_1, \pi_2, \ldots, \pi_n)$. Let $\gamma(\pi_i) = (V_i, E_i, v_{i,s}, v_{i,e}, \lambda_i, r_i)$, $1 \leq i \leq n$. By Lemma 1 we have

$$\begin{aligned} \mathsf{0\text{-}width}(\pi) &= \sum_{j=1}^{n} \mathsf{0\text{-}width}(\pi_j), \\ \mathsf{width}(\pi) &= \max_{j=1}^{n} (\mathsf{width}(\pi_j) + \sum_{i:1\leq i\leq n \wedge i\neq j} \mathsf{0\text{-}width}(\pi_i)). \end{aligned}$$

The latter equation can be rewritten as

$$\mathsf{width}(\pi) \;=\; \sum_{j=1}^{n} \mathsf{0\text{-}width}(\pi_j) \;+\; \max_{j=1}^{n} (\mathsf{width}(\pi_j) - \mathsf{0\text{-}width}(\pi_j)). \tag{23}$$

For each $i$ with $1 \leq i \leq n$, let us define $V[\pi_i, j] = \emptyset$ for every $j$ with $\mathsf{width}(\pi_i) < j \leq \mathsf{width}(\pi)$. We can then set

$$\begin{aligned} V[\pi, 1] &= V[\pi_1, 1] \cup \{v_s, v_e\}; \\ V[\pi, j] &= V[\pi_1, j], \quad \text{for } 2 \leq j \leq \mathsf{0\text{-}width}(\pi_1); \\ V[\pi, \mathsf{0\text{-}width}(\pi_1) + j] &= V[\pi_2, j], \quad \text{for } 1 \leq j \leq \mathsf{0\text{-}width}(\pi_2); \\ &\vdots \\ V[\pi, \textstyle\sum_{i=1}^{n-1} \mathsf{0\text{-}width}(\pi_i) + j] &= V[\pi_n, j], \quad \text{for } 1 \leq j \leq \mathsf{0\text{-}width}(\pi_n); \\ V[\pi, \textstyle\sum_{i=1}^{n} \mathsf{0\text{-}width}(\pi_i) + j] &= \cup_{i=1}^{n} V[\pi_i, \mathsf{0\text{-}width}(\pi_i) + j], \\ & \quad \text{for } 1 \leq j \leq \max_{j=1}^{n}(\mathsf{width}(\pi_j) - \mathsf{0\text{-}width}(\pi_j)). \end{aligned}$$

The sets $V[\pi, j]$ define a partition of $V$, since $V = (\cup_{i=1}^{n} V_i) \cup \{v_s, v_e\}$ and, for each $i$, the sets $V[\pi_i, j]$ define a partition of $V_i$ by the inductive hypothesis. We now show that such a partition satisfies both conditions in our statement.

Let $v_1$ and $v_2$ be distinct vertices in some $V[\pi, j]$, $1 \leq j \leq n$. We have already established in the proof of Lemma 1 that a cut $c$ in $\mathsf{cut}(\pi)$ either belongs to $\{v_s, v_e\}$ or else must have the form $c = c_1 \cdots c_n$ with $c_i \in \mathsf{cut}(\pi_i)$ for $1 \leq i \leq n$. As in Case 1, if either $v_1$ or $v_2$ belongs to the set $\{v_s, v_e\}$, then $v_1$ and $v_2$ cannot occur in the same cut in $\mathsf{cut}(\pi)$, since the only cuts in $\mathsf{cut}(\pi)$ with vertices in the set $\{v_s, v_e\}$ are $v_s$ and $v_e$. Consider now the case in which $v_1, v_2 \in \cup_{i=1}^{n} V_i$. We distinguish two subcases.

In the first subcase, there exists $i$ such that $v_1, v_2 \in V[\pi_i, j]$. If there exists a cut $c \in \mathsf{cut}(\pi)$ such that $v_1$ and $v_2$ both occur within $c$, then $v_1$ and $v_2$ must both occur within some $c' \in \mathsf{cut}(\pi_i)$. But this contradicts the inductive hypothesis on $\pi_i$.

In the second subcase, $v_1 \in V[\pi_{i'}, j']$ and $v_2 \in V[\pi_{i''}, j'']$, for distinct $i'$ and $i''$. Note that this can only happen if $\mathsf{0\text{-}width}(\pi) < j \leq \mathsf{width}(\pi)$, $\mathsf{0\text{-}width}(\pi_{i'}) < j' \leq \mathsf{width}(\pi_{i'})$ and $\mathsf{0\text{-}width}(\pi_{i''}) < j'' \leq \mathsf{width}(\pi_{i''})$, by our definition of the partition of $V$ and by (23). By the inductive hypothesis on $\pi_{i'}$ and $\pi_{i''}$, $v_1$ is not L-free in $\gamma(\pi_{i'})$ and $v_2$ is not L-free in $\gamma(\pi_{i''})$, which means that both $v_1$ and $v_2$ occur within the scope of some occurrence of the lock operator. Note however that $v_1$ and $v_2$ cannot occur within the scope of the same occurrence of the lock operator, since they belong to different subgraphs $\gamma(\pi_{i'})$ and $\gamma(\pi_{i''})$. Assume now that there exists a cut $c \in \mathsf{cut}(\pi)$ such that $v_1$ and $v_2$ both occur within $c$. This would be inconsistent with the definitions of $\Delta_\pi$ and cut (Definitions 4 and 5, respectively) since two vertices that are not L-free and that are not within the scope of the same occurrence of the lock operator cannot belong to the same cut.

Finally, it directly follows from the definition of our partition on $V$ and from the inductive hypothesis on the $\pi_i$ that all vertices in $V$ that are L-free in $\gamma(\pi)$ belong to some $V[\pi, j]$ with $1 \leq j \leq \mathsf{0\text{-}width}(\pi)$. This concludes the proof of our statement. □

The upper bound reported in Lemma 2 is tight. As an example, for any $i \geq 1$ and $k \geq 2$, let $\Sigma_{i,k} = \{a_1, \ldots, a_{i \cdot k}\}$. Consider now the class of IDL-expressions

$$\pi_{i,k} = \|(a_1 a_2 \cdots a_i, a_{i+1} a_{i+2} \cdots a_{2i}, \ldots, a_{i\cdot(k-1)+1} a_{i\cdot(k-1)+2} \cdots a_{i\cdot k}).$$

Let also $V_{i,k}$ be the vertex set of the IDL-graph $\gamma(\pi_{i,k})$. It is not difficult to see that $|V_{i,k}| = 2 \cdot i \cdot k + 2$, $\mathsf{width}(\pi_{i,k}) = k$ and

$$|\mathsf{cut}(\pi_{i,k})| = (2 \cdot i)^k + 2 \leq (2 \cdot i + \frac{2}{k})^k,$$

where the inequality results from our upper bound. The coarser upper bound presented before Lemma 2 would give instead $|\mathsf{cut}(\pi_{i,k})| < (2 \cdot i \cdot k + 2)^k$.

We can now turn to the discussion of the worst case running time for the algorithm in Figure 3. To simplify the presentation, let us ignore for the moment any term that solely depends on the input grammar $G$.

To store and retrieve items $[A \to \alpha \bullet \beta, c_1, c_2]$, $[a, c_1, c_2]$ and $[c_1, c_2]$ we exploit some data structure $T$ and access it using cut $c_1$ and cut $c_2$ as indices. In what follows we make the assumption that each access operation on $T$ can be carried out in an amount of time $\mathcal{O}(d(k))$, where $k = \mathsf{width}(\pi)$ and $d$ is some function that depends on the implementation of the data structure itself, to be discussed later. After we access $T$ with some pair $c_1$, $c_2$, an array is returned of length proportional to $|G|$. Thus, from such an array we can inquire in constant time whether a given item has already been constructed.

The worst case time complexity is dominated by the rules in Figure 3 that involve the maximum number of cuts, namely rules like (15) with three cuts each. The maximum number of different calls to these rules is then proportional to $|\mathsf{cut}(\pi)|^3$. Considering our assumptions on $T$, the total amount of time that is charged to the execution of all these rules is then $\mathcal{O}(d(k)\,|\mathsf{cut}(\pi)|^3)$. As in the case of the standard Earley algorithm, when the working grammar $G$ is taken into account we must include a factor of $|G|^2$, which can be reduced to $|G|$ using techniques discussed by Graham, Harrison, and Ruzzo (1980).

We also need to consider the amount of time required by the construction of relation $\Delta_\pi$, which happens on-the-fly, as already discussed. This takes place at Rules (16), (17) and (18). Recall that elements of relation $\Delta_\pi$ have the form $(c_1, X, c_2)$ with $c_1, c_2 \in \mathsf{cut}(\pi)$ and $X \in \Sigma \cup \{\varepsilon\}$. In what follows, we view $\Delta_\pi$ as a directed graph whose vertices are cuts, and thus refer to elements of such a relation as (labelled) arcs. When an arc in $\Delta_\pi$ emanating from a cut $c_1$ with label $X$ is visited for the first time, then we compute this arc and the reached cut, and cache them for possible later use. However, in case the reached cut $c_2$ already exists because we had previously visited an arc $(c_1', X', c_2)$, then we only cache the new arc. For each arc in $\Delta_\pi$, all the above can be easily carried out in time $\mathcal{O}(k)$, where $k = \mathsf{width}(\pi)$. Then the total time required by the on-the-fly construction of relation $\Delta_\pi$ is $\mathcal{O}(k\,|\Delta_\pi|)$. For later use, we now express this bound in terms of quantity $|\mathsf{cut}(\pi)|$. From the definition of $\Delta_\pi$ we can easily see that there can be no more than one arc between any two cuts, and therefore $|\Delta_\pi| \leq |\mathsf{cut}(\pi)|^2$. We obviously have $k \leq |V|$. Also, it is not difficult to prove that $|V| \leq |\mathsf{cut}(\pi)|$, using induction on the number of operator occurrences appearing within $\pi$. We thus conclude that, in the worst case, the total time required by the on-the-fly construction of relation $\Delta_\pi$ is $\mathcal{O}(|\mathsf{cut}(\pi)|^3)$.

From all of the above observations we can conclude that, in the worst case, the algorithm in Figure 3 takes an amount of time $\mathcal{O}(|G|\,d(k)\,|\mathsf{cut}(\pi)|^3)$. Using Lemma 2, we can then state the following theorem.

**Theorem 1** *Given a context-free grammar $G$ and an IDL-graph $\gamma(\pi)$ with vertex set $V$ and with $k = \mathsf{width}(\pi)$, the algorithm in Figure 3 runs in time $\mathcal{O}(|G|\,d(k)(\frac{|V|}{k})^{3k})$.*

We now more closely consider the choice of the data structure $T$ and the issue of its implementation. We discuss two possible solutions. Our first solution can be used when $|\mathsf{cut}(\pi)|$ is small enough so that we can store $|\mathsf{cut}(\pi)|^2$ pointers in the computer's random-access memory. In this case we can implement $T$ as a square array of pointers to sets of our parsing items. Each cut in $\mathsf{cut}(\pi)$ is then uniquely encoded by a non-negative integer, and such integers are used to access the array. This solution in practice comes down to the standard implementation of the Earley algorithm through a parse table, as presented by Graham et al. (1980). We then have $d(k) = \mathcal{O}(1)$ and our algorithm has time complexity $\mathcal{O}(|G|\,(\frac{|V|}{k})^{3k})$.

As a second solution, when $|\mathsf{cut}(\pi)|$ is quite large, we can implement $T$ as a trie (Gusfield, 1997). In this case each cut is treated as a string over set $V$, viewed as an alphabet, and we look up string $c_1\#c_2$ in $T$ ($\#$ is a symbol not in $V$) in order to retrieve all items involving cuts $c_1$ and $c_2$ that have been induced so far. We then obtain $d(k) = \mathcal{O}(k)$ and our algorithm has time complexity $\mathcal{O}(|G|\,k(\frac{|V|}{k})^{3k})$.

The first solution above is faster than the second one by a factor of $k$. However, the first solution has the obvious disadvantage of expensive space requirements, since not all pairs of cuts might correspond to some grammar constituent, and the array $T$ can be very sparse in practice. It should also be observed that, in the natural language processing applications discussed in the introduction, $k$ can be quite small, say three or four.

To conclude this section, we compare the time complexity of CFG parsing as traditionally defined for strings and the time complexity of parsing for IDL-graphs. As reference for string parsing we take the Earley algorithm, which has already been presented in Section 6. By a minor change proposed by Graham et al. (1980), the Earley algorithm can be improved to have time complexity $\mathcal{O}(|G| \cdot n^3)$, where $G$ is the input CFG and $n$ is the length of the input string. We observe that, if we ignore the factor $d(k)$ in the time complexity of IDL-graph parsing (Theorem 1), the two upper bounds become very similar, with function $(\frac{|V|}{k})^k$ in IDL-graph parsing replacing the input sentence length $n$ from the Earley algorithm.

We observe that function $(\frac{|V|}{k})^k$ can be taken as a measure of the complexity of the internal structure of the input IDL-expression. More specifically, assume that no precedence constraints at all are given for the words of the input IDL-expression. We then obtain IDL-expressions with occurrences of the I operator only, with a worst case of $k = \frac{|V|}{2} - 1$.

Then $\mathcal{O}((\frac{|V|}{k})^k)$ can be written as $\mathcal{O}(c^{|V|})$ for some constant $c > 1$, resulting in exponential running time for our algorithm. This comes at no surprise, since the problem at hand then becomes the problem of recognition of a bag of words with a CFG, which is known to be NP-complete (Brew, 1992), as already discussed in Section 2.

Conversely, no I operator may be used in the IDL-expression $\pi$, and thus the resulting representation matches a finite automaton or word lattice. In this case we have $k = 1$ and function $(\frac{|V|}{k})^k$ becomes $|V|$. The resulting running time is then a cubic function of the input length, as in the case of the Earley algorithm. The fact that (cyclic or acyclic) finite automata can be parsed in cubic time is also a well-known result (Bar-Hillel et al., 1964; van Noord, 1995).

It is noteworthy to observe that in applications where $k$ can be assumed to be bounded, our algorithm still runs in polynomial time. As already discussed, in practical applications of natural language generation, only few subexpressions from $\pi$ will be processed simultaneously, with $k$ being typically, say, three or four. In this case our algorithm behaves in a way that is much closer to traditional string parsing than to bag parsing.

We conclude that the class of IDL-expressions provides a flexible representation for bags of words with precedence constraints, with solutions in the range between pure word bags without precedence constraints and word lattices, depending on the value of $\mathsf{width}(\pi)$. We have also proved a fine-grained result on the time complexity of the CFG parsing problem for IDL-expressions, again depending on values of the parameter $\mathsf{width}(\pi)$.

## 8. Final Remarks

Recent proposals view natural language surface generation as a multi-phase process where finite but very large sets of candidate sentences are first generated on the basis of some input conceptual structure, and then filtered using statistical knowledge. In such architectures, it is crucial that the adopted representation for the set of candidate sentences is very compact, and at the same time that the representation can be parsed in polynomial time.

We have proposed IDL-expressions as a solution to the above problem. IDL-expressions combine features that were considered only in isolation before. In contrast to existing formalisms, interaction of these features provides enough flexibility to encode strings in cases where only partial knowledge is available about word order, whereas the parsing process remains polynomial in practical cases.

The recognition algorithm we have presented for IDL-expressions can be easily extended to a parsing algorithm, using standard representations of parse forests that can be extracted from the constructed parse table (Lang, 1994). Furthermore, if the productions of the CFG at hand are weighted, to express preferences among derivations, it is easy to extract a parse with the highest weight, adapting standard Viterbi search techniques as used in traditional string parsing (Viterbi, 1967; Teitelbaum, 1973).

Although we have only considered the parsing problem for CFGs, one may also parse IDL-expressions with language models based on finite automata, including $n$-gram models. Since finite automata can be represented as right-linear context-free grammars, the algorithm in Figure 3 is still applicable.

Apart from natural language generation, IDL-expressions are useful wherever uncertainty on word or constituent order is to be represented at the level of syntax and has to be linearized for the purpose of parsing. As already discussed in the introduction, this is an active research topic both in generative linguistics and in natural language parsing, and has given rise to several paradigms, most importantly immediate dominance and linear precedence parsing (Gazdar, Klein, Pullum, & Sag, 1985), discontinuous parsing Daniels and Meurers (2002), Ramsay (1999), Suhre (1999) and grammar linearization (Götz & Penn, 1997; Götz & Meurers, 1995; Manandhar, 1995). Nederhof, Satta, and Shieber (2003) use IDL-expressions to define a new rewriting formalism, based on context-free grammars with IDL-expressions in the right-hand sides of productions. By means of this formalism, fine-grained results were proven on immediate dominance and linear precedence parsing.[5]

IDL-expressions are similar in spirit to formalisms developed in the programming language literature for the representation of the semantics of concurrent programs. More specifically, so called series-parallel partially ordered multisets, or series-parallel pomsets, have been proposed by Gischer (1988) to represent choice and parallelism among processes. However, the basic idea of a lock operator is absent from series-parallel pomsets.

## Acknowledgments

A preliminary version of this paper has appeared in the *Proceedings of the 7th Conference on Formal Grammars* (FG2002), Trento, Italy. The notions of IDL-graph and cut, central to the present study, are not found in the earlier paper. We wish to thank Michael Daniels, Irene Langkilde, Owen Rambow and Stuart Shieber for very helpful discussions related to the topics in this paper. We are also grateful to the anonymous reviewers for helpful comments and pointers to relevant literature. The first author was supported by the PIONIER Project *Algorithms for Linguistic Processing*, funded by NWO (Dutch Organization for Scientific Research). The second author was supported by MIUR under project PRIN No. 2003091149_005.

5. In the cited work, the lock operator was ignored, as it did not affect the weak generative capacity nor the compactness of grammars.